\documentclass{llncs}
\usepackage[pdftex]{graphicx}
\usepackage{amsmath,amsfonts,amssymb}
\usepackage{multirow} 
\usepackage{rotating}
\usepackage{floatrow}
\usepackage{subfigure}
\usepackage{color}
\usepackage{longtable}
\usepackage{tikz}
\definecolor{myred}{rgb}{.8,.0,.0}

\newcommand\blfootnote[1]{%
  \begingroup
  \renewcommand\thefootnote{}\footnote{#1}%
  \addtocounter{footnote}{-1}%
  \endgroup
}
\begin{document}
\pagestyle{headings}  
\title{Hands-Free Segmentation of Medical Volumes via Binary Inputs}
\author{Florian Dubost\inst{1,2} \and Loic Peter\inst{1} \and
Christian Rupprecht\inst{1,3} \and Benjamin~Gutierrez~Becker\inst{1} \and Nassir Navab\inst{1,3}}
%
\institute{Computer Aided Medical Procedures, Technische Universit\"{a}t M\"{u}nchen, Germany
\and
Biomedical Imaging Group Rotterdam, Erasmus MC, Rotterdam, The Netherlands
\and
Computer Aided Medical Procedures, Johns Hopkins University, Baltimore, USA \\
}
\maketitle
\begin{abstract}
We propose a novel hands-free method to interactively segment 3D medical volumes. In our scenario, a human user progressively segments an organ by answering a series of questions of the form ``\textit{Is this voxel inside the object to segment?}''. At each iteration, the chosen question is defined as the one halving a set of candidate segmentations given the answered questions. For a quick and efficient exploration, these segmentations are sampled according to the Metropolis-Hastings algorithm. Our sampling technique relies on a combination of relaxed shape prior, learnt probability map and consistency with previous answers. We demonstrate the potential of our strategy on a prostate segmentation MRI dataset. Through the study of failure cases with synthetic examples, we demonstrate the adaptation potential of our method. We also show that our method outperforms two intuitive baselines: one based on random questions, the other one being the thresholded probability map. 
\end{abstract}
\section{Introduction} \blfootnote{The final publication is available at Springer via http:://dx.doi.org/[xxx]}
The segmentation of medical images or volumes is a key research topic in medical image analysis. The segmentation of objects of interest - e.g. organs or tumors - is a key process for operation planning, navigation or design of personalized prosthesis. Interactive segmentation is often a well-suited framework as it allows the user to actively participate in the segmentation process and correct possible mistakes or refine the segmentation. However this interactive aspect can rise issues when the segmentation has to be made during surgery: (i) the process of zooming and navigating through slices can be overwhelming and time-consuming, (ii) the hands of the clinicians are already busy with the operation itself. The use of hands-free techniques can thus be handy and is in general appreciated by clinicians \cite{liu,miller} as they significantly reduce the labelling effort for medical data.

In many popular methods for interactive segmentation the user gives indications - scribbles, bounding boxes - as an input to the algorithm \cite{grad,grab}. Once the indications are given, the algorithm runs autonomously without new input from the user. A example of a hands-free technique in this framework would be Eyegaze \cite{eye} which is based on eye tracking. However this technique still involves much navigation and zooming and needs a calibration. 

Another way to perform interactive segmentation is to build an algorithm which iteratively includes the indications of the user, following a refinement technique. The simplest way to handle this is to display the resulting segmentation after each interaction. Each input of the user will then be seen as a hard constraint \cite{gauriau}. Another general idea of this framework is to use the answers already provided by the user to hint for areas of high uncertainty and guide the user in the search. One possible way to locate such areas is through segmentation sampling. State of the art methods of segmentation sampling can be based on Markov Chain Monte Carlo (MCMC) \cite{tu,rupp} or Gaussian Process \cite{le}. Both methods \cite{rupp,le} proved to be effective in 2D but encounter - because of the use of Geodesic Distance Transform - high running time when performed on 3D data. 

\begin{figure}[t]
\centering
\includegraphics[height=6cm]{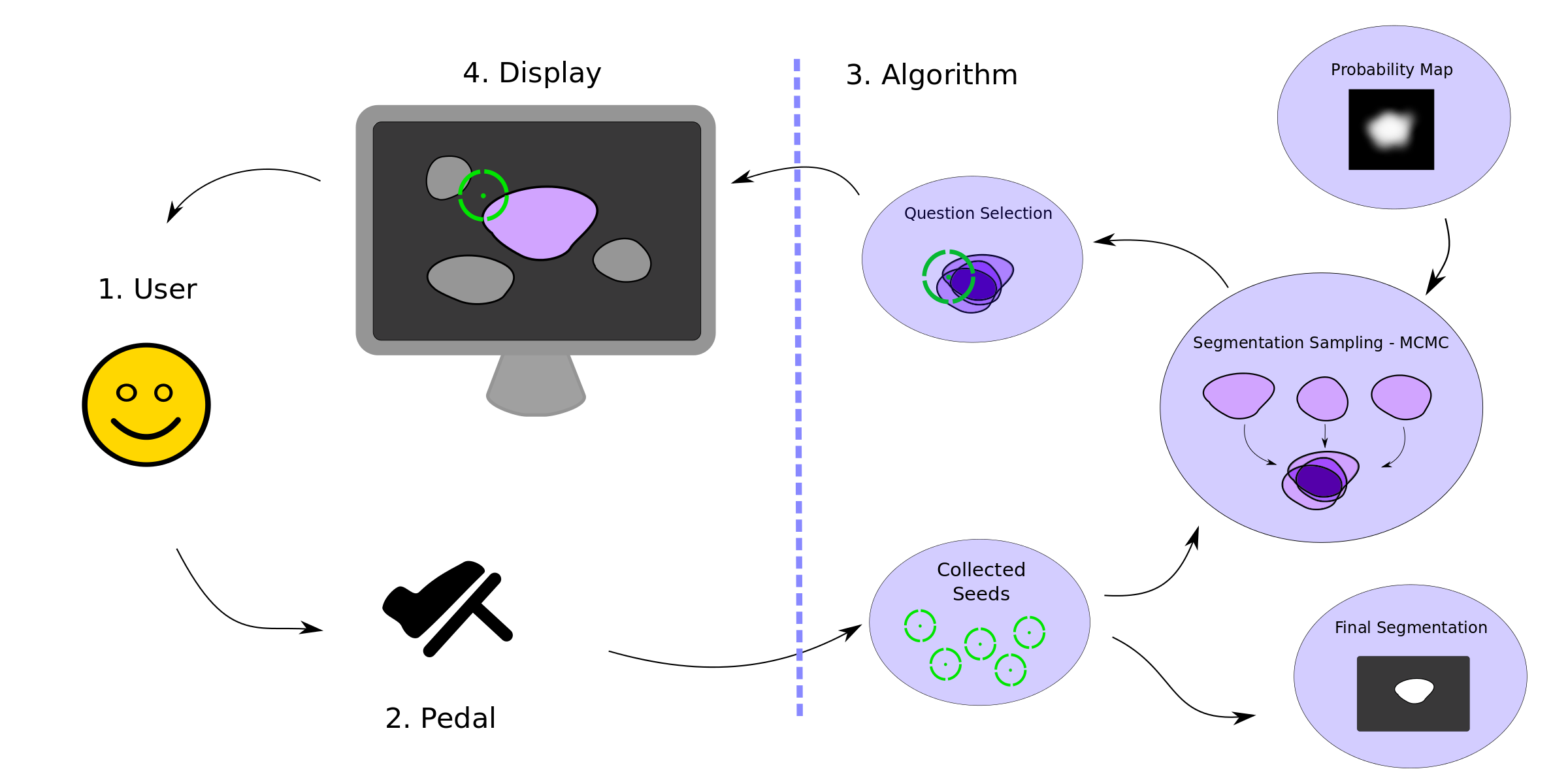}
\caption{\textbf{Diagram summarizing the processes of our method.} The user (1) communicates for instance via a pedal (2) with the algorithm (3) which outputs question voxels to the user (4). Using a probability map learnt offline (Sec. \ref{sec:probaMap}) and previous answers from the user, the algorithm samples several segmentations (Sec. \ref{sec:mcmc}) and finds the area where they disagree the most. The question voxel is taken within this area (Sec. \ref{sec:qVoxel}). The question is of the form ``\textit{Is this voxel inside the object to segment?}''. The answer of the user provides a seed which halves the space of candidates segmentations. The final segmentation is computed from the set of seeds provided by the user and by running a last segmentation sampling procedure (Sec. \ref{sec:finalSeg}).}
\label{fig:diag}
\end{figure}

In this paper, we propose a novel hands-free interactive segmentation method. In our scenario a human user segments an object of interest from a 3D medical volume only by answering questions of the type ``\textit{Is this voxel inside the object to segment?}''. These answers are binary interactions - "yes"/"no" - and can be easily recorded trough a pedal or voice recognition system. They provide a set of positive and negative seeds to compute the final segmentation. In order to choose the question voxels we sample candidate segmentations thanks to a MCMC framework. This sampling process relies on an adaptive weighting between a probability map learnt off-line and the consistency with previous answers. If the probability map is misleading, the algorithm detects it and changes accordingly.
The answer of the user halves then the space of the sampled candidate segmentations, following a dichotomic search in this space.
We propose a diagram (Fig.~\ref{fig:diag}) summarizing our technique. We evaluated the performance of our method on a 3D MRI prostate segmentation dataset. Through the study of failure cases generated with synthetic examples, we demonstrate the adaptation potential of our method. Our results demonstrate that our technique can correct inaccurate annotations or ameliorate imprecise ones in a reasonable time.
\section{Methods}
In the following paragraphs we start by briefly explaining the learning of the probability map (Sec. \ref{sec:probaMap}). In the next section we detail the core of our method and contribution for the segmentation sampling of the MCMC technique (Sec. \ref{sec:mcmc}). Our idea consists in combining a relaxed shape prior, a learnt probability map and the consistency with previous answers. One of our main contributions is the adaptation capability of our algorithm, which can identify misleading probability maps and adapt accordingly. The last paragraphs briefly review how to propose questions voxels from the sampled segmentations (Sec. \ref{sec:qVoxel}) and how to compute the final segmentation of the algorithm, once all $K$ questions have been answered (Sec. \ref{sec:finalSeg}).

Let $\Gamma = \{1,...,H\}\times\{1,...,W\}\times\{1,...,D\}$ be a three-dimensional lattice and $V$ a volume defined on $\Gamma$. We call $S$ the space of segmentations, i.e. the set of functions $\textbf{s}:\Gamma\mapsto\{0,1\}$. If the voxel $v(x,y,z)$ is inside the segmented object then $\textbf{s}(x,y,z)=1$, otherwise  $\textbf{s}(x,y,z)=0$.
\subsection{Probability Map}
\label{sec:probaMap}
Our method uses as prior knowledge a probability map $\pi$ defined over $\Gamma$. This probability map is obtained with a classifier trained offline. $\pi(v)$ is an estimation of the probability that the voxel $v(x,y,z)$ belongs to the targeted object. We have no prior information on the quality of this probability map. 

To obtain $\pi$, we use an AdaBoost classifier \cite{boost} based on Haar features \cite{viola}, which we more precisely defined and sampled as in \cite{feat}.
We denote the stumps $h_{t}$ for $t=\{1,...,T\}$, where $T$ is the number of boosting iterations. We compute the decision function $H$ as the sum of the $h_{t}$.
In order to rescale the output values so that $0 < \pi(x,y,z) < 1$ we apply a sigmoid function to the score $H$ \cite{sig}.
\subsection{MCMC Framework}
\label{sec:mcmc}
We would like to generate segmentations to approximate the space of probable segmentations and then use the answer of the user to halve this space, following a dichotomic search.
In this section we present our technique to sample candidate segmentations. We follow the MCMC framework proposed and used in \hbox{\cite{tu,rupp}}. The idea is to generate segmentations by running through a Markov Chain. We define the Markov Chain over a state space $\mathcal{X}$ so that from a state $\textbf{x} \in \mathcal{X}$ we can compute a unique segmentation $\textbf{s}(\textbf{x})$. The states are parametrized with transformation coefficients based on a shape prior (see next paragraph). 

The process goes as follows: from a current state $\textbf{x}$, we induce small variations using a proposal distribution $Q$ to generate a new proposed state $\textbf{x'}$. We can then compute the likelihood of the new underlying segmentation $\textbf{s}(\textbf{x'})$ using a posterior probability $P$. The new state $\textbf{x'}$ is accepted with a transition probability $\alpha$ defined as
\begin{equation}
\alpha(\textbf{x'}|\textbf{x}) = \min\Bigl\{1,\frac{P(\textbf{x'})Q(\textbf{x}| \textbf{x'})} {P(\textbf{x})Q(\textbf{x'}| \textbf{x})}\Bigl\}.
\end{equation}
If the move if accepted, the proposed state becomes the current one and we reiterate the process. Otherwise we come back to $\textbf{x}$ and a new state is proposed.
\subsubsection{Parametrization of Segmentations}
\label{sec:shape}
The objective is here to explain how segmentations are represented. We decided to use shape models for it allows us to generate 3D segmentations with a very low running time. Following a similar idea than in \cite{crem} we define a relaxed notion of shape based on signed distance functions. Given a training set of $m$ relaxed shapes $\mathcal{Y} = \{y_{1},...,y_{m}\}$, we can calculate the mean $\mu$ and the $n$ first eigenmodes $\psi_{1},...,\psi_{n}$. To create a new relaxed shape we compute
\begin{equation}
y = \mu + \sum_{i = 1}^{n} b_{i} \psi_{i},
\end{equation}
where $b_{1}, ... , b_{n}$ are the eigencoefficients of the shape prior. To widen the space of segmentations we allow as well resizing and rigid transformations such as translation and rotation. 
Therefore a state $\textbf{x}$ is defined by $7 + n$ parameters, as $\textbf{x} = (a, t_{x}, t_{y}, t_{z}, \alpha, \beta, \gamma, b_{1}, ... , b_{n})$, where $a$ is the size parameter, $t_{x}, t_{y}$ and $t_{z}$ translation parameters, $\alpha, \beta$ and $\gamma$ rotation parameters and $b_{1}, ... , b_{n}$ the eigencoefficients of the shape prior.
The resulting segmentation $\textbf{s}(\textbf{x}) \in S$ is computed as $y(\textbf{x})$ thresholded at 0.
\subsubsection{Posterior Probability}
This probability is encoding how likely a state $\textbf{x}$ - and its underlying segmentation $\textbf{s(x)}$ - is, given the already provided answers $\Sigma$ and the probability map $\pi$. We denote it as $P(\textbf{x}|\Sigma)$ and compute it as
\begin{equation}
\label{posterior}
P(\textbf{x}|\Sigma) \propto \frac{1}{1 - L(\textbf{x}) + \beta_{k} g(\textbf{x})},
\end{equation}
where $L(\textbf{x})$ denotes the likelihood between the probability map $\pi$ and the proposed segmentation $\textbf{s}(\textbf{x})$, $g(\textbf{x})$ is a penalty term including the $k$ previous answers from the user, and $\beta_{k}$ a weighting parameter between these two objectives after $k$ questions. By doing so, we consider as likelier the segmentations that are close to the probability map and compatible with the user responses. The relative weighting $\beta_{k}$ of these two terms is adjusted after each question by checking the compatibility of the posterior with the provided answers. Thereby, if the posterior probability is mistaken, its impact is gradually decreasing. The next paragraphs expose our model for $L$, $\beta_{k}$ and $g$. 
\paragraph{Likelihood - Probability Map.}
\label{sec:likelihood}
To evaluate whether a candidate segmentation is close to the probability map, we use a maximum likelihood scheme. To simplify the following notations, we write $v(x,y,z) = v(t)$ where $t$ is a parameter spanning the whole volume.
For a given voxel $v(t)$ we assume $\textbf{s}(t)$ follows a Bernoulli distribution $\mathrm B(\pi(t))$. If we consider $\textbf{s}(1),...,\textbf{s}(|\Gamma|)$ iid samples, the weighted log-likelihood is given by
\begin{equation}
L(\textbf{x}) = \frac{1}{|\Gamma|}\log(\mathcal{L}(S = \textbf{s} | \pi)) = \frac{1}{|\Gamma|}\sum\limits_{t \in \Gamma} \textbf{s}(t)\log(\pi (t)) + (1 - \textbf{s}(t))\log(1 - \pi(t)).
\end{equation}
This quantity is always negative and reaches its maximum - $L(\textbf{x}) = 0$ - when perfect match occurs.
\paragraph{Penalty Term.}
\label{sec:penalty}
We introduce a penalty term $g(\textbf{x})$ to include the information provided by the $k$ previous answers of the user in the estimation of the posterior probability $P$. This way, we would like to penalize a candidate segmentation $\textbf{s(\textbf{x})}$ which is not compatible with the given answers. We model the answers as a seed location $\sigma = \{x_{\sigma}, y_{\sigma}, z_{\sigma}\}$ and a corresponding label $a(\sigma) \in \{0,1\}$. 
We denote $\Sigma_{err}$ the set of $m$ seeds violated by the candidate segmentation, with $m \leq k$. We consider that a segmentation violates a seed $\sigma$ when its prediction for this seed does not match the label provided by the user $a(\sigma)$.\\
Following the definition of signed distance functions, $|y(\textbf{x},\sigma_{err})|$ gives a measure of the distance between the violated seed $\sigma_{err}$ and the border of the proposed segmentation $\textbf{s}(\textbf{x})$. We compute therefore the penalty term as
\begin{equation}
g(\textbf{x}) = \sum_{\sigma \in \Sigma_{err}} |y(\textbf{x},\sigma)|.
\end{equation}
\paragraph{Adaptive Weighting Parameter.}
For the weighting parameter $\beta_{k}$ between the two objective functions $L$ and $g$ we propose an automatic adaptable setting.
The idea consists in updating $\beta$ at each question to progressively verify whether the probability map $\pi$ can be trusted and adapt the loss function $- L(\textbf{x}) + \beta g(\textbf{x})$ accordingly.
If the probability map is accurate, $\beta$ should stay close to $0$, otherwise beta should increase. The setting is inspired from online transfer learning \cite{online}.
$\beta$ is initialized to $\beta_{0}$ and a new value $\beta_{k+1}$ is computed after each question $k$ according to
\begin{equation}
\beta_{k+1} = \max(\beta_{max},\beta_{k}*e^{-\epsilon \mu l(1/2,\pi(\sigma))}),
\end{equation}
where $\mu$ is a parameter encoding the amplitude of the update, i.e. the learning rate, $\beta_{max}$ a parameter encoding the maximum value for beta to avoid divergence, $\epsilon$ is the agreement between the answer of the user and the probability map and $l$ a loss function  encoding the confidence of the probability map in its prediction. In our case we chose $l(x,y) = |x - y|$ to measure the distance between the neutral answer $1/2$ and the probability $\pi(\sigma)$. The closer to $1/2$ the probability $\pi(\sigma)$ is, the less it influences the update of $\beta$.

Our definition for $\epsilon$ is led by the one of the Dice similarity coefficient. We do not consider true negative seeds informative. Let $\pi_{threshold}$ be the probability map thresholded at $0.5$. We set $\epsilon = 1$ if $\pi_{threshold}(\sigma) =  a(\sigma) = 1$; $\epsilon = -1$ if $\pi_{threshold}(\sigma) \neq  a(\sigma)$ and ($\pi_{threshold}(\sigma) = 1$ or $a(\sigma) = 1$); and $\epsilon = 0$ if $\pi_{threshold}(\sigma) =  a(\sigma) = 0$ which is considered as uninformative and therefore does not update the value of beta.\\
\begin{figure}
\centering
\includegraphics[height=3.0cm]{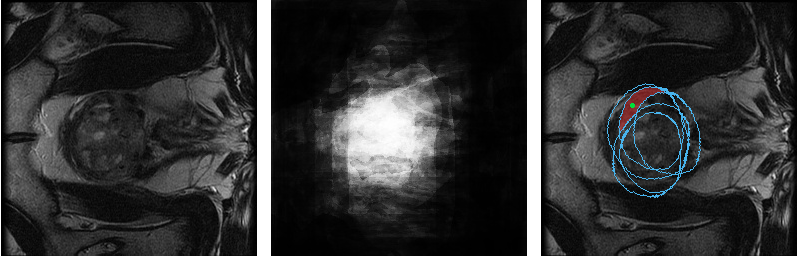}
\caption{\textbf{Illustration of the steps of our algorithm on a MRI image of the prostate.} From left to right: original image; probability map obtained from boosting; overlapping of the candidate segmentations for the question selection during the MCMC. The question voxel (green) is taken on the centroid of the selected region (red). (Source of the original image: Prostate Segmentation Challenge MICCAI09)}
\label{fig:qSelection}
\end{figure}
\subsection{Question Voxel}
\label{sec:qVoxel}
In order to compute the question voxel from the sampled segmentations, we follow the same framework as \cite{rupp}. By superposing the accepted sampled segmentations, we divide the volume into several regions. We choose the voxel question as the centroid of the most unsure of these regions (Fig. \ref{fig:qSelection}).
\subsection{Final Segmentation}
\label{sec:finalSeg}
After $K$ question have been asked and the $K$ corresponding seeds have been collected, we now compute the final segmentation $\textbf{s}_{f}$.
We sample candidate segmentations reusing the MCMC framework and compute their posterior probability $P(\textbf{x})$ according to equation (\ref{posterior}). During this step the weighting parameter is fixed to $\beta_{K}$, i.e. the lastly updated $\beta_{k}$. The final segmentation is taken as the one maximizing $P(\textbf{x})$.
\section{Experiments}

Our experimental evaluation was performed on the dataset of the Prostate Segmentation Challenge MICCAI09. This dataset is a collection of 15 3D MRI annotated images coming each from a different patient. The voxel resolution is $0.55 \times 0.55 \times 5 mm^{3}$. The images have an average voxel size of $256\times 256\times 32$. We used the T2-weighted images for our experiments.
\subsection{Experimental Settings}
\label{sec:param}
We follow a 5-fold cross-validation framework, where the training set is used to learn the probability maps and shape models. To generate new shapes, we retain only the $n=3$ first eigenmodes of the shape, which defines our state space $\mathcal{X}$ with $10$ dimensions.
Concerning the weighting parameter $\beta_{k}$, we set $\beta_{0} = 1$, $\mu = 3$ and $\beta_{max} = 4$. During the MCMC we perform a burn-in step of 100 iterations and run 25 iterations between each sampled segmentation. The total number of sampled segmentations at each question is $N = 15$. During the exploration of the states $\textbf{x} \in \mathcal{X}$ in the segmentation sampling, the proposal distribution $Q$ draws the parameters of $\textbf{x}$ from Gaussian distributions centered on their current value.
We use the Dice Similarity Coefficient (DSC) \cite{dice} to evaluate the performance of our algorithm.
We implemented our algorithms in C++ and ran the experiments on a Intel i7-4702MQ 2.20GHz CPU.
The computation time between each question is low enough to allow an interactive use of the algorithm. We performed an experiment to study the time statistics over the dataset. Over $165$ questions - $45$ per patient - the computation time between two questions was in average $4.2s$, in median $3.9s$ and had a standard deviation of $1s$.
\subsection{Results}
\subsubsection{Synthetic Probability Map}
\begin{figure}
\centering
\includegraphics[height=4cm]{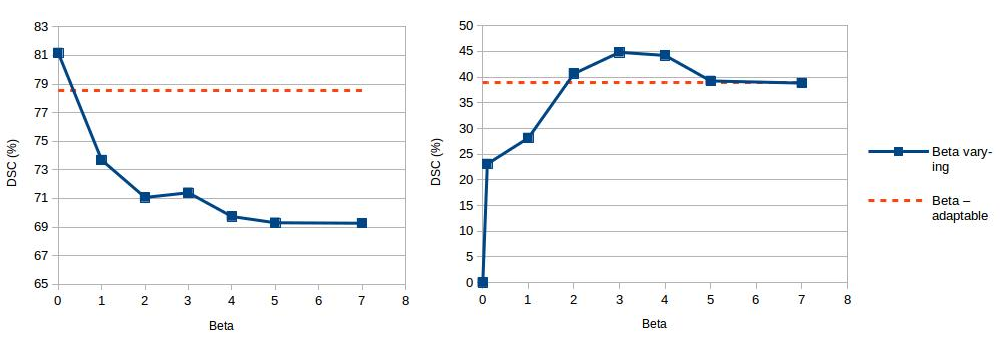}
\caption{\textbf{Evolution of performance of our algorithm on synthetic data with different values of a fixed $\beta$.} The straight line shows the performance obtained using the automatic setting of $\beta$. On the left, we use the blurred ground truths as probability maps (1). We notice that if the probability maps are already performing well, the answers of the user do not increase the performance. This  can be detected in a very few questions looking at the automatic setting of beta. The segmentation can then be considered as already too accurate to be improved by our algorithm. The DSC is capped to $81\%$ because of the lack of freedom of our shape model. On the right, we use misleading probability maps (2). We notice that increasing beta correlates with a significantly better performance in this scenario. Note that $\beta$ has much more influence over the performance in this case than in (1). Here our algorithm learns to identify and ignore inaccurate probability maps.}
\label{fig:gt_20q}
\end{figure}

In our first experiments we demonstrate the adaptation capability of our method through the automatic setting of parameter $\beta_{k}$. Instead of using the learnt probability maps we create synthetic ones to cover the two extreme case scenarios: (1) the probability map is almost perfect and can be trusted, (2) the probability map is inaccurate and shouldn't be considered to generate segmentations. To simulated these probability maps, we use for (1) the blurred ground truth and for (2) the translated blurred ground truth such that the dice overlap with the original ground truth is zero. In Fig. \ref{fig:gt_20q} we plot respectively for (1) and (2) curves showing the evolution of the dice similarity coefficient (DSC) according to a manual setting of $\beta$ ranging from $0$ to $7$. On the same plot we show the result obtained using the automatic adaptable setting of beta detailed in section \ref{sec:mcmc}. 
\begin{figure}
\centering
\includegraphics[height=5cm]{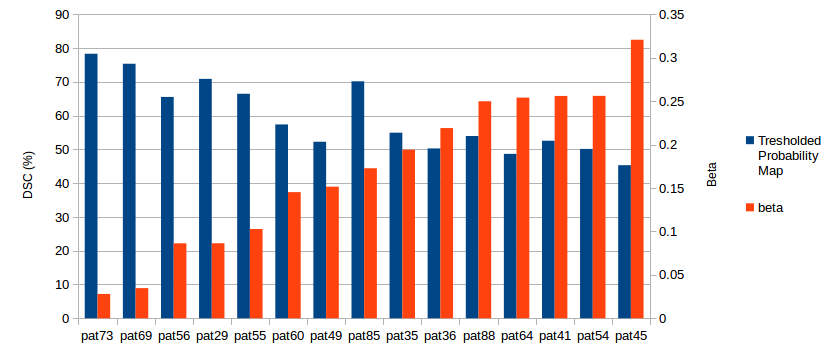}
\caption{\textbf{Automatic setting of $\beta$ after 30 questions in comparison the Dice score of the tresholded probability map.} The results are displayed for each patient individually. In this experiment we use the learnt probability maps. As expected, we notice a trend of the coefficient $\beta$ to adapt to the quality of the probability maps. Low beta for trustworthy ones, high beta for the ones of poorer quality. This fits to the expected behaviour of the coefficient $\beta$.}
\label{fig:evbeta}
\end{figure}
\subsubsection{Learnt Probability Map}
\begin{figure}
\centering
\includegraphics[height=5.5cm]{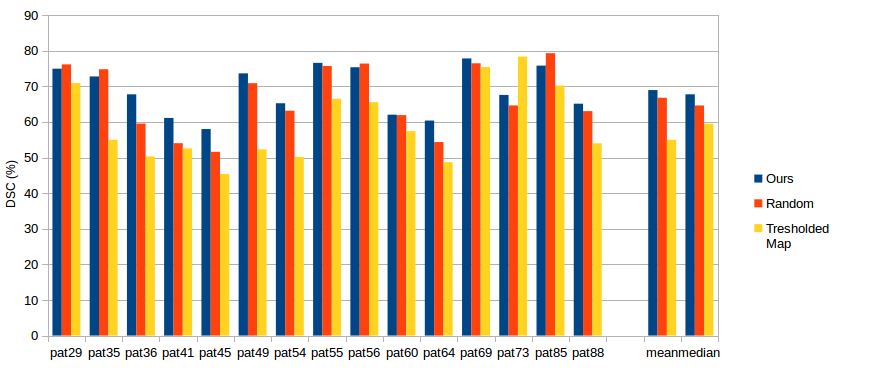}
\caption{\textbf{Comparison of Dice Scores on the Prostate Dataset.} Comparison between our method (blue) and two baselines: random questions (red) and the thresholded probability map (yellow). The last two columns are the mean and median over patients. As pictured in Fig. \ref{fig:gt_20q} the use of shape models bounds the DSC to $80\%$ in average.}
\label{fig:results}
\end{figure}

To assess the quality of our segmentations we compute the DSC after $30$ questions. We compare our technique with two intuitive baselines: the first one corresponds to probability map from boosting thresholded at $0.5$. The second one consists in asking the questions at random voxels instead of trying to find the most unsure area with the MCMC framework. The results are shown in Fig.~\ref{fig:evbeta}, \ref{fig:results}. 
If we look more closely, we notice that our algorithm performs better than the random questions baseline for the patients for which the probability map performed the worst. This fits well our motivation to retrieve poor segmentations. However we notice for instance that for patient 73, the thresholded probability baseline performs better than both our method and the random questions baseline. This could be resulting from a lack of freedom of our shape-model which therefore impede the mimic of unusual shapes as the one in patient 73.
The algorithm proposed by \cite{rupp} cannot be applied here because 3D GDT is not feasible in real time.
Dowling et al. \cite{dowling} report results on the same dataset and have more heterogeneous results. Our initial model - the probability map - is in average not as accurate as theirs and we expect better results if this component is improved via the use of more sophisticated learning techniques. However, our contribution here is mainly to illustrate the interactive scenario with a restriction to binary inputs and our initial model has not been optimized for this specific task. We also believe that there is room for more accurate shape models on this dataset, since the number of training volumes for this task was limited here.
\section{Conclusion}
We presented an interactive hands-free method to segment objects of interest in medical volumes. Experiments demonstrate the potential of our method to retrieve inaccurate and misleading segmentations.
Using a probability map and a shape prior we are able to locate informative areas to ask questions. The use of shape models to generate segmentations allows a quick computational time between each question. We provided an automatic adaptable setting for weighting the influence of the probability map.
This method could be useful in surgery, to allow for instance last minute corrections of incorrect segmentations.
Future work could include interactive updates of the probability map with the answers of the user, combining it for instance with an unsupervised model.
%
%
%

\end{document}